\documentclass[conference]{IEEEtran}
\IEEEoverridecommandlockouts

\usepackage{microtype}
\usepackage{graphicx}
\usepackage{subfigure}
\usepackage{booktabs} 

\usepackage{hyperref}

\usepackage{amsmath}
\usepackage{amssymb}
\usepackage{mathtools}
\usepackage{amsthm}
\usepackage{cancel}

\usepackage[capitalize,noabbrev]{cleveref}

\theoremstyle{plain}

\theoremstyle{definition}

\theoremstyle{remark}

\usepackage[textsize=tiny]{todonotes}

\usepackage{hyperref}
\usepackage{url}
\usepackage{graphicx}

\usepackage[english]{babel}
\usepackage{amsthm}

\usepackage{algorithm}
\usepackage{algorithmic}

\usepackage{wrapfig}
\usepackage{multicol}

\title{Concurrent Credit Assignment for Data-efficient Reinforcement Learning \thanks{This work was supported by the CNRS and French ANR AgileNeuRobot [grant no 
ANR-20-CE23-002].}}

\author{Emmanuel Daucé\\
Institut de Neurosciences de la Timone\\
Centrale Marseille, CNRS\\
Marseille, France\\
\texttt{emmanuel.dauce@centrale-marseille.fr}
}

\begin{document}

\maketitle

\begin{abstract}
The capability to widely sample the state and action spaces is a key ingredient toward building effective reinforcement learning algorithms. The variational optimization principles exposed in this paper emphasize the importance of an occupancy model to synthesizes the general distribution of the agent's environmental states over which it can act (defining a virtual ``territory''). The occupancy model is the subject of frequent updates as the exploration progresses and that new states are undisclosed during the course of the training. By making a uniform prior assumption, the resulting objective expresses a balance between two concurrent tendencies, namely the widening of the occupancy space and the maximization of the rewards, reminding of the classical exploration/exploitation trade-off. Implemented on an actor-critic off-policy on classic continuous action benchmarks, it is shown to provide significant increase in the sampling efficacy, that is reflected in a reduced training time and higher returns, in both the dense and the sparse rewards cases. 
\end{abstract}

\section{Problem statement}

Online learning in the real world implies dealing with very large, potentially unlimited environments, over which the data to collect is seemingly infinite. 
\emph{Efficient} exploration is thus one of the key aspects of open-ended learning \cite{10.3389/fnbot.2019.00115}, when no final model of the environment can feasibly be expected to be engineered or trained. 
On the one hand, having access to unlimited data is very beneficial for the training of complex multi-layered perceptrons, for they are known to rely on large datasets to improve their performance. 
On the other hand, the circular dependence between the learning algorithm and the data on which it operates renders the 
learning very tricky, at high risk of data overfitting and trapping in local optima.
The open-ended learning problem is generally addressed through the lens of the reinforcement learning framework \cite{sutton1998introduction}, where rewards are collected during the interaction, and the selection of action is fit so as to maximize the total number of positive rewards, and prevent the encounter of negative ones. Fitting behaviour to rewards is however at the risk of ignoring important data from the rest of the environment, where putatively more rewarding regions may be neglected. The agreement of reward-seeking (that is exploitation) with data collecting (that is exploration), is still one of the fundamental issues of modern artificial intelligence.

An important effort has recently been put on reframing the reinforcement learning setup into a more general probabilistic inference framework, allowing to link rewards seeking and data modelling under a single perspective \cite{furmston2010variational,levine2018reinforcement,haarnoja2018soft,abdolmaleki2018maximum,fellows2019virel}. This greater focus over the data collection problem conducts to the development of learning algorithms containing some forms of exploration bonuses, including ``curiosity'' drives \cite{schmidhuber2009simple,pathak2017curiosity}, intrinsic rewards \cite{oudeyer2007intrinsic} and pseudo-counts \cite{bellemare2016unifying,tang2017exploration}.
However, if optimizing on rewards alone comes with a well-defined Bellman optimum, there is still no consensus about the objective followed when optimizing both the rewards and the data collection \cite{eysenbach2019if}. The data collection problem is effectively shadowed by the reward maximization objective, and is still considered as a side component of the learning procedure.
An important body of work has recently been devoted to addressing the data collection problem as such, with the notable design of the MaxEnt algorithm \cite{hazan2019provably} and State Marginal Matching \cite{lee2019efficient}  that  aim at fitting the distribution of the states encountered to a uniform distribution, in the absence of definite rewards. This is here referred as a MaxEnt-on-state principle (or MaxEnt to be short), not to be confounded with the MaxEnt-on-actions principle implemented in the soft actor critic \cite{haarnoja2018soft} for instance. Following a MaxEnt objective means optimizing the policy so as the states visited are maximally variable, ideally following a uniform distribution. 
We develop in the following a possible generalization of the MaxEnt principle, that brings a considerable simplification in the expression  of the evidence lower bound (ELBO) with regards with the existing literature \cite{furmston2010variational,abdolmaleki2018maximum,fellows2019virel}. In contrast to pure MaxEnt, our approach provides a way to combine the MaxEnt objective with a reward maximization objective, under a variational inference perspective. 
This gives ways toward optimizing the policy with respect to the distribution of the data, and provides a principled justification to the use of intrinsic rewards in the design of reinforcement learning algorithms.


%
%

\section{Principles}

\subsection{Density matching RL}
Consider an agent acting in a fully observable environment. The state of the environment is given by the observation variable (or vector) $s\in \mathcal{S}$, with $\mathcal{S}$ the set of all possible states. The agent exerts a control on the environment through its actuators. Let $a \in \mathcal{A}$ a command sent to the actuators, with $ \mathcal{A}$ the set of all motor commands. 
The interaction of the agent with its environment is organized as follows: at each time step $t$, a data sample $s_t \in \mathcal{S}$ is observed, and an action $a_t$ is emitted. 
It is supposed here, for simplicity, that the dynamics of the environment is Markovian, i.e. the state observed at time $t$ only depends on the previous observation $s_{t-1}$ and previous action $a_{t-1}$. Moreover, the environment provides at each time step a scalar value called the reward, i.e. $r_t \in \mathbb{R}$.

The decision of which action to choose relies on a \emph{policy} $\pi$, that maps the current observation to the action space. For generality, we consider the policy as a conditional probability $\pi(A=a|S=s)$ (or $\pi(a|s)$ in short). The capital letters $S$ and $A$ are then random variables on $\mathcal{S}$ and $\mathcal{A}$, and the lower cases $s$ and $a$ are some specific realizations of the corresponding random variables. 
The series of states visited during the interaction of the agent with its environment forms a sequence $\tau=(s_0, ..., s_t, ...)$, with $s_0$ the initial observation. During this visit, a certain number of rewards are collected, and $R(\tau)$ is the total (discounted) return obtained over $\tau$, i.e. $R(\tau) = \sum_t \gamma^t r_t$, with $\gamma \in [0,1[$ a discounting factor that sums up the rewards up to an ``horizon'' of the order of $\frac{1}{1-\gamma}$.

The learning objective we address in this paper is the the ``density matching'' (or probability matching) objective \cite{sabes1996reinforcement}. 
A reward should indicate in which proportion the different states (and actions) should be visited (and selected) during trials (with the idea that the states providing high return should be visited more often than the ones providing low returns). Solving the reinforcement learning problem then means to match the external cue to an actual distribution of visit over states and actions, where a differential in rewards only indicates a difference in the number of visits, allowing to seek rewards in a flexible way (so it is also referred as to ``soft'' reinforcement learning \cite{haarnoja2018soft}). This idea stems back from empirical observations on human and animal behaviors, and was coined the ``matching law'' in the operant conditioning literature \cite{herrnstein1961relative,eysenbach2019if}.  

In short, the density matching framework considers the reinforcement learning in the terms a control problem. Let $s$ be the current observation, $a$ the action chosen, $\tau$ the issuing trajectory. Consider each trajectory $\tau$ as the realization of a certain conditional distribution $q_\pi(\tau|s,a)$, that depends on the current policy $\pi$.
Note $Q_\pi(s,a) = \mathbb{E}_{\tau\sim q_\pi(\tau|s,a)} R(\tau)$ the expected discounted sum of rewards collected after $(s,a)$. The function $Q$, that maps $\mathcal{S}\times\mathcal{A}$ to $\mathbb{R}$, is known as the action-value or ``critic''. In our specific probabilistic setup, $Q_\pi(s,a)$ is moreover interpreted as a log probability $\hat{\pi}(a|s)\propto \exp(Q_\pi(s,a))$ that reflects the frequency at which a current action $a$ should be selected in the context $s$. The general objective of the density matching optimization is:
\begin{align}\label{eq:PM-objective}
&\min_{\pi \in \Pi} \mathbb{E}_{s} \mathcal{D}_\text{KL}(\pi(.|s)|| \hat{\pi}(.|s))\nonumber\\
&\equiv \min_{\pi \in \Pi} \mathbb{E}_{s,a,\tau} \log \pi(a|s) - R(\tau)
\end{align}
The equivalence of the first and the second objective comes from the specific probabilistic design of $\hat{\pi}(a|s)$.
Importantly, this objective does not coincide with the mainstream Bellman objective \cite{bellman1966dynamic}, that is to find:
\begin{align}\label{eq:Bellman-objective}
\max_{\pi \in \Pi} \mathbb{E}_{s,a,\tau} R(\tau)
\end{align}
Both are however strongly consistent,
referred as the ``soft'' versus ''hard'' formulations of the problem.  
The main difference is that the soft optimization is expected promote multimodal responses and allow for a better exploration of the different options at stake \cite{haarnoja2018soft}. Moreover, the optimum of the soft optimization problem is generally a non-deterministic policy.

\subsection{Sample efficacy and the exploration objective}

One important component of the soft optimization problem (eq. \ref{eq:PM-objective}) is the maximum entropy $\max_{\pi \in \Pi} -\mathbb{E}_{s,a} \log \pi(a|s)$ on actions that conduct to select more varied action samples during the optimization. Specifically, in the absence of rewards, it consists to select the actions uniformly. One can remark, however \cite{o2020making}, that such a uniform selection on actions is not equivalent to a uniform exploration of the possible states of the environment. Indeed, in the absence of rewards, the maximal sample efficacy is attained when the trajectories uniformly cover the state space. This objective is known as the maximum entropy \emph{on states} objective \cite{hazan2019provably,lee2019efficient}. 

Take any trajectory $\tau$ sampled from the controlled environment and select any $s\in \tau$, then the maximum entropy objective assumes that the distribution of $s$ is uniform over the state space. This objective distribution is noted $p^*(s)$.   
Take now any $s'\in\tau$ and 
and $\rho_\pi(s')$ the marginal over all 
$\tau$'s. The maximum entropy on states objective can now be formally defined as: $\max_{\pi \in \Pi} -\mathbb{E}_{s'\sim \rho_\pi(s')} \log \rho_\pi(s')$.
More generally, taking $p^*(s)$ a uniform distribution as a baseline, the maximum entropy objective can be framed as a density matching objective like:
\begin{align}\label{eq:successor PM}
\max_{\pi \in \Pi} \mathbb{E}_{s'} - \log \rho_\pi(s') + \log p^*(s')
\end{align}
This objective is known as the State Marginal Matching (SMM) in the literature \cite{lee2019efficient}. 
The SMM objective is not supposed equal the policy matching (eq. \ref{eq:PM-objective}) or even the Bellman (eq. \ref{eq:Bellman-objective}) objectives. 
Following this exploration objective requires the use of a \emph{model} of the controlled system. For instance, one may build a data model $\tilde{\rho}_\pi$ from sampling states from the visited trajectories, 
from an explicit occurrence count in the discrete case, or with a parametric or non-parametric estimator. The density of visits over the state space reflects more generally the state domain over which the data is collected, 
that is used in return to implement the control policy.

It is classical, in the literature, to interpret the SMM objective (\ref{eq:successor PM}) as an \emph{intrinsic reward} attached to the state $s'$, that can take the form of a state count in the discrete case \cite{bellemare2016unifying}, or a non-parametric model in the continuous case \cite{lee2019efficient}. In the last case, the total reward used in the optimization is $r(s,a,s') \equiv r(s,a) - \log \rho_\pi(s')$, that is the sum of an extrinsic and an intrinsic reward, that reflects the ``surprise'' of visiting $s'$ after $(s,a)$.  This construction needs to be considered cautiously, for this sort of intrinsic reward is non stationary with regards to policy updates. On contrary, each update of the policy modifies in return the marginal distribution $\rho_\pi$, which may conduct to instabilities and cycles in the optimization process.
In consequence, the intrinsic rewards are implemented in the form of an ``exploration bonus'', that is supposed to vanish during the course of the training, without clear convergence guarantees. 

\subsection{Occupancy models}

Let us first consider in more detail the data model $\rho_\pi$, constructed from trajectory samples of the controlled environment.  This model reflects a complex process of data generation that starts with observing a state $s$, choosing an action $a$, and then observing both a trajectory $\tau$ and a return $R(\tau)$. 
The state data $\{..., s',...\}$ is collected through the different trajectories visited during this process. Let $\rho_\pi$ be the actual trajectory distribution, as it is sampled from the environment, and $\tilde{\rho}_\pi$ a statistical model constructed from observing the data. Both distributions are non stationary, i.e. vary according to the policy updates.
We thus need a way to express more precisely this dependency, and then include the changes of $\rho$ in a more general optimization objective.

\subsubsection{State occupancy}
Dating back from \cite{dayan1993improving}, an occupancy distribution is a distribution on states, designed so as to match with the distribution measured over the trajectories of the MDP. 
Importantly, it ignores the time order at which the different states are visited.
For instance, following the definitions of \cite{puterman2014markov, ho2016generative, hazan2019provably}, a $\gamma$-absorbing state occupancy of a Markov Decision process (with a policy $\pi$) is the (discounted) density of visit of the states 
of the environment when starting from the initial distribution $p(S_0)$.
It is defined, as:  
\begin{align} \label{eq:occupancy}
			\rho_\pi(s) = (1-\gamma) p_0(s) + \gamma \sum_{s',a'} p(s|s',a') \pi(a'|s') \rho_\pi(s') 
\end{align}
so that any policy $\pi$ settled on an MDP defines an occupancy on the states of that MDP. 


%
\subsubsection{Conditional occupancy}
Following the same reasoning, let $\rho_\pi(S'|s,a)$ the \emph{conditional occupancy} be defined recursively, as the occupancy on the states $s'$ that follow $(s,a)$ in the iteration of the dynamics. 
This conditional distribution provides a description of the ``future'' of $s$, that is the distribution of states that will most probably follow $s$. It can be seen as an instance of the ``successor'' representation of states initially proposed by \cite{dayan1993improving}, with $s'$ being the state measured  ``further away in time''.


\subsubsection{Total return estimator}
One can then construct an estimator of the total discounted return (that is : the discounted sum of rewards observed over the whole trajectory) from observing a single state sample over this trajectory.  
Consider for instance the series of rewards encountered when following $\tau$. It comes that $\forall s' \in \tau$,  $\tilde{R}(s')=\mathbb{E}_{a'\sim\pi(A|s')}\frac{r(s',a')}{1-\gamma}$ is an estimator of $R(\tau)$, so that: 
\begin{align}
	Q(s_t,a_t) &= \mathbb{E}_{\tau:s_t, a_t\in\tau} \sum_{t'>t} \gamma^{(t'-t)} r_{t'} \nonumber\\
	&\approx  \mathbb{E}_{\substack{s' \sim \rho_\pi(S'|s_t,a_t)\\a'\sim\pi(A|s')}} \sum_{t'>t} \gamma^{(t'-t)} r(s',a') \nonumber\\
	&=  \mathbb{E}_{\substack{s' \sim \rho_\pi(S'|s_t,a_t)\\a'\sim\pi(A|s')}} \frac{r(s',a')}{1-\gamma} \nonumber\\
	&\triangleq \mathbb{E}_{s' \sim \rho_\pi(S'|s_t,a_t)} \tilde{R}(s')
\end{align} 


\subsection{Evidence Lower Bound}

We now come to the main argument: in the general case, the sample efficacy and the reward maximization objectives are considered separately from each other. We assume here that both optimizations need to be done \emph{concurrently} 
in order to attain a final control policy that explicitly implements a trade-off between the exploration and the exploitation objectives. We thus develop in the following an original optimization objective that implements this concurrent optimization in a principled way.

The main probabilistic model to consider for the optimization is the policy, $\pi_\theta(.|s)$, that is a conditional probability, i.e. a stochastic mapping from the observation space to the action space. The parameters of the model are optimized according to the loss function considered, e.g. from stochastic gradient  descent if the mapping is differentiable. 
Let $\theta$ be the parameters of the policy. Those parameters can be seen as the realization of a policy model $\theta\sim p(\Theta)$. 
Examples of policy parameters are the look-up tables $\theta=\{Q(s,a)\}_{s,a}$ in the discrete case. The intervention of $\pi_\theta$ in the data generation process provokes a dependence, so that $\rho_\pi$ now writes $\rho(.|\theta)$ and $\tilde{\rho}_\pi$ writes $\tilde{\rho}(.|\theta)$.

The policy model $p(\Theta)$ is unknown, but it can be approached by the following approximate \emph{evidence lower bound} (ELBO):
\begin{align}\label{eq:ELBO}
\log p(\theta) \geq \mathbb{E}_{s'\sim \rho(.|\theta)} \log p(\theta|s') - \log \tilde{\rho}(s'|\theta) + \log p^*(s')
\end{align}
with $p(\theta|s')$ being the likelihood of the parameters of the policy an $p^*(s')$ being a prior distribution on the trajectories of the controlled environment.
Finding the most likely parameters of the policy now relies on a \emph{dual} objective,  that is both maximizing the likelihood of the parameters and maximizing the fitting of the data with the prior. The two objectives are connected by a common sampling $\rho(.|\theta)$, and the non-stationarity of the data distribution becomes an explicit \emph{result} of the optimization on $\rho$.

This ELBO synthesizes the main elements of this concurrent optimization process. On the one hand, the likelihood of the parameters is involved, directly or indirectly, by the measure of the rewards obtained over the different trajectory samples, reflecting the reward maximization objective (Bellman objective). 
On the other hand, the matching of the posterior $\tilde{\rho}$ with a prior $p^*$ is consistent with the pursuit of a specific control objective.
This provides a way, for instance, to inject the MaxEnt principle in the equation (that is maximizing the sampling efficacy), in which case $p^*$ is a uniform distribution on the trajectories, and would play the same role than the Gaussian prior in variational auto-encoders \cite{kingma2013auto}.
This ELBO is \emph{approximate} because the true distribution $\rho_\pi$ that is the actual sampling of $s'$ under $\pi$ is different from the data model $\tilde{\rho}$ that is constructed from a mere observation of those samples. The accuracy of the ELBO is thus dependent on the quality of the model and the diversity of the data actually sampled from $\pi$. 

The idea to express the reinforcement learning problems in terms of variational optimization is not new, and has been the subject of an abundant literature \cite{furmston2010variational,levine2018reinforcement,haarnoja2018soft,abdolmaleki2018maximum,fellows2019virel}, in which the learner aims at predicting  a latent objective from its interactions with the environment. One recent example is for instance inferring a ``task'' (or even a ``sub-task'') from non-rewarded interactions with the environment, in a pre-training phase \cite{eysenbach2018diversity}. Another body of work considers instead an objective of maximizing a (behavioral) mutual information between a policy and a model \cite{mohamed2015variational,houthooft2016vime} etc. A neat difference here is the fact that $s'$ (the ``data''), that takes the place of the latent variable, is not inferred from $\theta$, but merely sampled from the environment (and the policy). This conducts to the fact that our variational optimization setup is \emph{non-inferential}, on contrary to the mainstream variational inference. This sort of non-inferential/variational presentation of the training objective can be found, to some extent, in the stochastic control literature \cite{levine2018reinforcement}. 

%


\subsection{Implementation}

We have seen that the optimization should rely on two main components, namely (i) an occupancy model $\tilde{\rho}$ and (ii) a parametric policy $\pi_\theta$, both being distributions of probabilities, 
with the data model being optimized from observing the data samples.
The data consists of all the past observations, that can be seen as a set of trajectories $\mathcal{D} = \{\tau, \tau', \tau'', ...\}$, and the state occupancy model $\tilde{\rho}_\mathcal{D}$ is a distribution of probability, that can be constructed explicitly from the transition probabilities, or estimated directly from the data.
The empirical occupancy distribution $\tilde{\rho}_\mathcal{D}$ 
synthesizes the general distribution of the agent's environmental states over which it can act (defining a virtual ``territory''). The occupancy model is the subject of frequent updates as the exploration progresses and that new states are undisclosed during the course of the training.

With a parametric policy and a state occupancy model in hand, one can now implement a stochastic gradient descent over the parameters of the policy. Consider first the look-up table case where $\mathcal{S}$ and $\mathcal{A}$ are discrete and $\theta=\{Q(s,a)\}_{s,a}$ is a table of parameters.
Consider each observation $(s,a)$ as a sample. From each $(s,a)$, one can record a trajectory $\tau$, read $s'\in \tau$ and obtain a return $\tilde{R}(s')$, making a total data sample $(s,a,s')$.
Assume now that:
\begin{itemize}
    \item $\pi_\theta(a|s)\propto\exp(\beta Q(s,a))$ (Softmax decision with $\beta$ the ``inverse temperature''), with $Q(s,a)$ the critic, and assume $\theta(s,a)=Q(s,a)$.
    \item The likelihood of $\theta$, that is sampled independently at each $(s,a)$, is supposed to follow a normal law $\mathbb{E}_{s,a,s'} p(\theta|s')=\mathbb{E}_{s,a,s'}\mathcal{N}(\theta(s,a)| \tilde{R}(s'),\frac{1}{\lambda})$ with $\lambda$ a fixed ``precision'' parameter.
\end{itemize}
Then, each sample $(s,a,s')$ provides an ELBO sample:
\begin{align}\label{eq:ELBO}
\text{ELBO} &= \mathbb{E}_{s,a,s'}\text{ELBO}(s,a,s')\nonumber\\
     \equiv \mathbb{E}_{s,a,s'} &-\frac{\lambda}{2} (Q(s,a)-\tilde{R}(s'))^2 - \log \tilde{\rho}_\mathcal{D}(s') + \log p^*(s')
\end{align}


\subsection{Optimization}
Then a policy gradient ascent \cite{williams1992simple} on $Q$ can be implemented over these samples. 
Assuming ${p}^*(s)$ a constant for simplicity, the policy gradient measured at $(s,a)$ is:
\begin{align}\label{eq:ELBO-update}
\nabla_Q &\mathbb{E}_{s,a,s'}\text{ELBO}(s,a,s') \equiv \mathbb{E}_{s,a,s'} \lambda (\tilde{R}(s')-Q(s,a)) \nonumber\\
&+ \frac{1}{\beta}\nabla_{\beta Q(s,.)} \log \pi_\theta(a|s) \text{ELBO}(s,a,s')
\end{align} 

The gradient estimator interestingly combines two terms that reflect the two main (and concurrent) influences of the ELBO objective. The first term, that is the gradient of the likelihood with regards to $\theta$, is the classical ``reward prediction error'' update of the action-value found in the majority of RL algorithms. The second term interestingly reflects an ``amortized'' policy gradient, that conducts to treat $- \log \tilde{\rho}(s')$ as an intrinsic reward, with a correction term that moderates the update for it may not deviate too much from the current estimate.
Finally, in the case of the classic Monte Carlo sampling of trajectories, this conducts to repeat the following steps:
\begin{enumerate}
	\item run an experiment with $\pi$
	\item add the trajectory $\tau$ to the dataset $\mathcal{D}$
	\item take some samples $s,a,s'\in\tau$, and do policy gradient ascent on  $\text{ELBO}(s,a,s')$
	\item (periodically) update $\tilde{\rho}_\mathcal{D}$ with the new data
\end{enumerate}

\subsection{Online setup and Bellman recurrence}

In the online case, a last element of implementation 
is considering the Bellman recurrence \cite{sutton1988learning}, that rests on estimating the cumulative $R(\tau)$ with the sum of the current reward $r_t$ plus its best ($\gamma$-discounted) proximal estimate.
In our case, two separate estimators $Q_\text{R}$ and $Q_\text{S}$ need to be set up, with $Q_\text{R}$ estimating  the total return and $Q_\text{S}$ estimating $- \mathbb{E}_{s'}\log \tilde{\rho}(s')$ over the full trajectory, with:
\begin{align}
&Q_\text{R}(s_t,a_t) \simeq r(s_t,a_t) + \gamma Q_\text{R}(s_{t+1},a_{t+1})  \\
&Q_\text{S}(s_t,a_t) \simeq -(1-\gamma) \log \tilde{\rho}_\mathcal{D}(s_{t+1}) + \gamma Q_\text{S}(s_{t+1},a_{t+1})\\
&\text{ELBO}(s_t,a_t) \simeq -\frac{\lambda}{2} (Q(s_t,a_t)-Q_\text{R}(s_t,a_t))^2 + Q_\text{S}(s_t,a_t)
\label{eq:approx_ELBO}\end{align}
Then $\mathbb{E}_{s'}\tilde{R}(s')$ is `replaced'' by  $Q_\text{R}(s,a)$ and $\mathbb{E}_{s'}-\log\tilde{\rho}_\mathcal{D}(s')$ is ``replaced'' by $Q_\text{S}(s,a)$ in the ELBO formulas. Each state-action value function implements a distinct objective, with $Q_\text{R}(s,a)$ heading toward maximizing the reward, and $Q_\text{S}(s,a)$ heading toward the maximization of the entropy of the occupancy. Each objective being followed concurrently, our method was named the ``Concurrent Credit Assignment'' (CCA) method.

\section{Simple example}

\begin{figure}[t]
\centerline{
    \includegraphics[width=\linewidth]{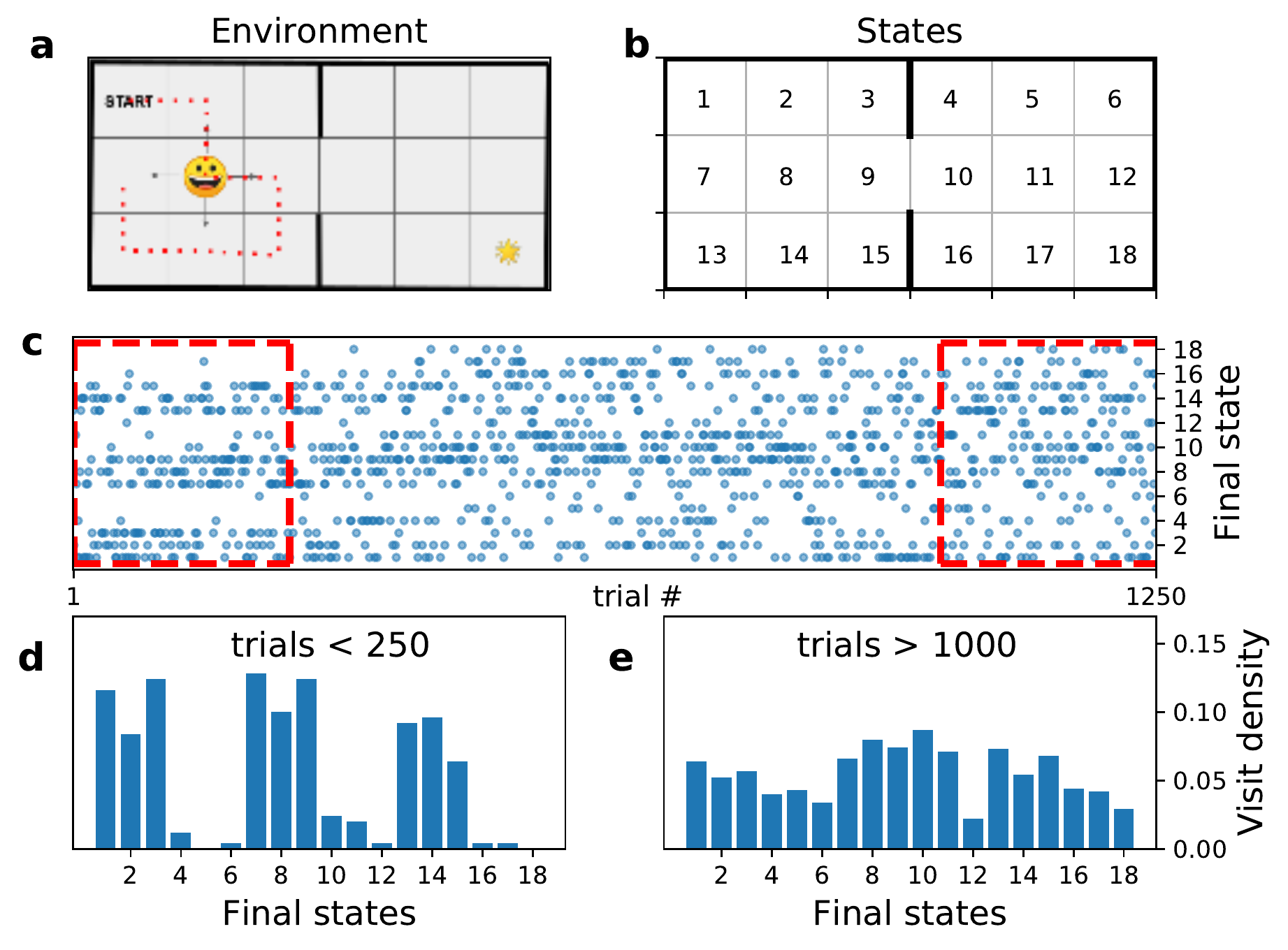}
}
\caption{{\bf Task 1 : reward-agnostic setting ($\lambda=0$)} {\bf a.} A two-rooms environment. Starting from the upper-left corner, the agent has to execute a sequence of 7 elementary actions $a_1,...,a_7$, each elementary action being in (E,S,W,N). The reward is equal to 1 if the final state is the lower-right corner ($s'=18$), and 0 otherwise. {\bf b.} Environment states. {\bf c.} Series of final states attained during the first 1250 trials. {\bf d.} Empirical distribution of final states for $\text{\#trial}<250$ (warm-up phase). {\bf d.} Empirical distribution of final states for $t>1000$ (permanent regime). Discrete CCA algorithm. $\alpha=0.3$, $\beta = 1$,  $\eta=0.005$.}\label{fig:explore}
\end{figure}

Let us now consider a simple setup that illustrates the method. 
Let the environment be a grid world with 18 discrete states organized in two rooms, with a single transition allowing to pass from room A to room B (see figure \ref{fig:explore}a). Let $\mathcal{A}$ a control space accounting for a single degree of freedom (here a discrete set of actions i.e. $\mathcal{A}=\{E,S,W,N\}$).
Importantly, the agent always starts in the upper left corner of room A, and can only execute 7 moves. 
The final state $s' = s_7$ attained after the 7 moves 
reflects, to some point, the result,  or the outcome, of the control. Moreover, a reward of 1 is given when the agent reaches the lower right corner, and 0 otherwise.
By construction, 
when acting at random under this 7-steps path length constraint, the chance to end-up in the first room is significantly higher than the chance to end up in the second room.
The environment is indeed constructed in such a way that some final states (for instance upper right and lower right corners) are very unlikely to be reached when acting in a purely random manner. 


The agent has to train a policy $\pi$ and an occupancy model $\tilde{\rho}(s')$, with 
$s'$ the final state obtained at the end of the control.
There are two task at hand. A first task is a simple exploration task and the objective is to uniformly sample the final state space. A second task consists in maximizing the total rewards obtained. 
From a practical standpoint, the training requires to find the parameters $\theta = \{ Q_\theta(s,a)\}_{s\in\mathcal{S}, a\in\mathcal{A}}$.  Let $p^*(s')=\frac{1}{|\mathcal{S}|}$ be a constant with $|\mathcal{S}|$ the cardinal of $\mathcal{S}$.
Let $n_{s'}$ be the number of times that the final state $s'$ was reached from the beginning of the learning session, and $\tilde{\rho}(s')=\frac{n_{s'}}{\sum_{u'}n_{u'}}$ the empirical density of visit. In this specific implementation, the occupancy model is updated in a ``leaky'' way, with a leak parameter $\eta$, in order to adapt faster to the changes occurring in the distribution (see algorithm \ref{algo:E3D}). The general idea is to adapt the occupancy model a little more ``faster'' than the policy, for the policy gradient to follow the distribution changes as they are recorded in the model.  The update procedure is provided in algorithm \ref{algo:E3D}. 

When the reward is not considered (that is $\lambda=0$), the ELBO drive contributes to progressively ``expand'' the visiting territory, with any peripheral state attained increasing the probability to reach its further neighbors, recursively, up to the final limit of the state space. In small environment like the one proposed here, the limit is rapidly attained and an approximate uniform alternation of visits is observed over the full state space. 
The final distribution of states is compared in figure \ref{fig:explore}c between the initial phase and the permanent regime attained after approximately 1000 trials. In the starting phase, a strong bias in favor of the first room is observed (states 1-3;7-9;13-15, see figure \ref{fig:explore}d). 
In contrast, a time consistent uniform pattern of visit is observed in the permanent regime, that illustrates the capability of the CCA method to construct policies specifically devoted to the wide exploration of the environment  (figure \ref{fig:explore}e).     

\begin{algorithm}[t]
	\caption{Discrete Concurrent Credit Assignment}\label{algo:E3D}
	\begin{algorithmic}
		\REQUIRE{$\alpha$, $\beta$, $\eta$, $\lambda$}
		\STATE $Q \leftarrow \vec{0}_{|\mathcal{S}|\times|\mathcal{A}|}$
		\STATE $\rho \leftarrow$ Uniform
		\WHILE{number of trials not exceeded}
			\STATE sample $\tau_{1:n} \sim \pi$ 
			\STATE read $s_n,R(s_n)$
			\STATE $R_\text{S} = - \log \tilde{\rho}_\mathcal{D}(s_n)$
			\STATE $\rho \leftarrow (1-\eta) \rho + \eta \mathbf{1}_{S=s_n} $ 
			\FOR{$i \in 1..n-1$}
			    \STATE ${Q_\text{R}(s_i,a_i) \leftarrow (1 - \alpha) Q_\text{R}(s_i, a_i) + \alpha R(s_n)}$
			    \STATE  ${Q_\text{S}(s_i,a_i) \leftarrow (1-\alpha) Q_\text{S}(s_i, a_i) + \alpha R_\text{S}}$
			    \STATE $Q(s_i,a_i) \leftarrow (1-\alpha\lambda) Q(s_i, a_i) + \alpha \lambda Q_\text{R}(s_i,a_i)$
			    \STATE $Q(s_i,a_i) \leftarrow Q(s_i,a_i)+ \frac{\alpha}{\beta} \text{\sc elbo}(s_i,a_i)$
			    \FOR {$a \in \mathcal{A}$}
			        \STATE $Q(s_i,a) \leftarrow Q(s_i,a)-\frac{\alpha\pi_\theta(a|s_i)}{\beta}  \text{\sc elbo}(s_i,a_i)$
			    \ENDFOR
			\ENDFOR
		\ENDWHILE
	\end{algorithmic}
\end{algorithm}

\begin{figure}
		\centerline{\includegraphics[width=.9\linewidth]{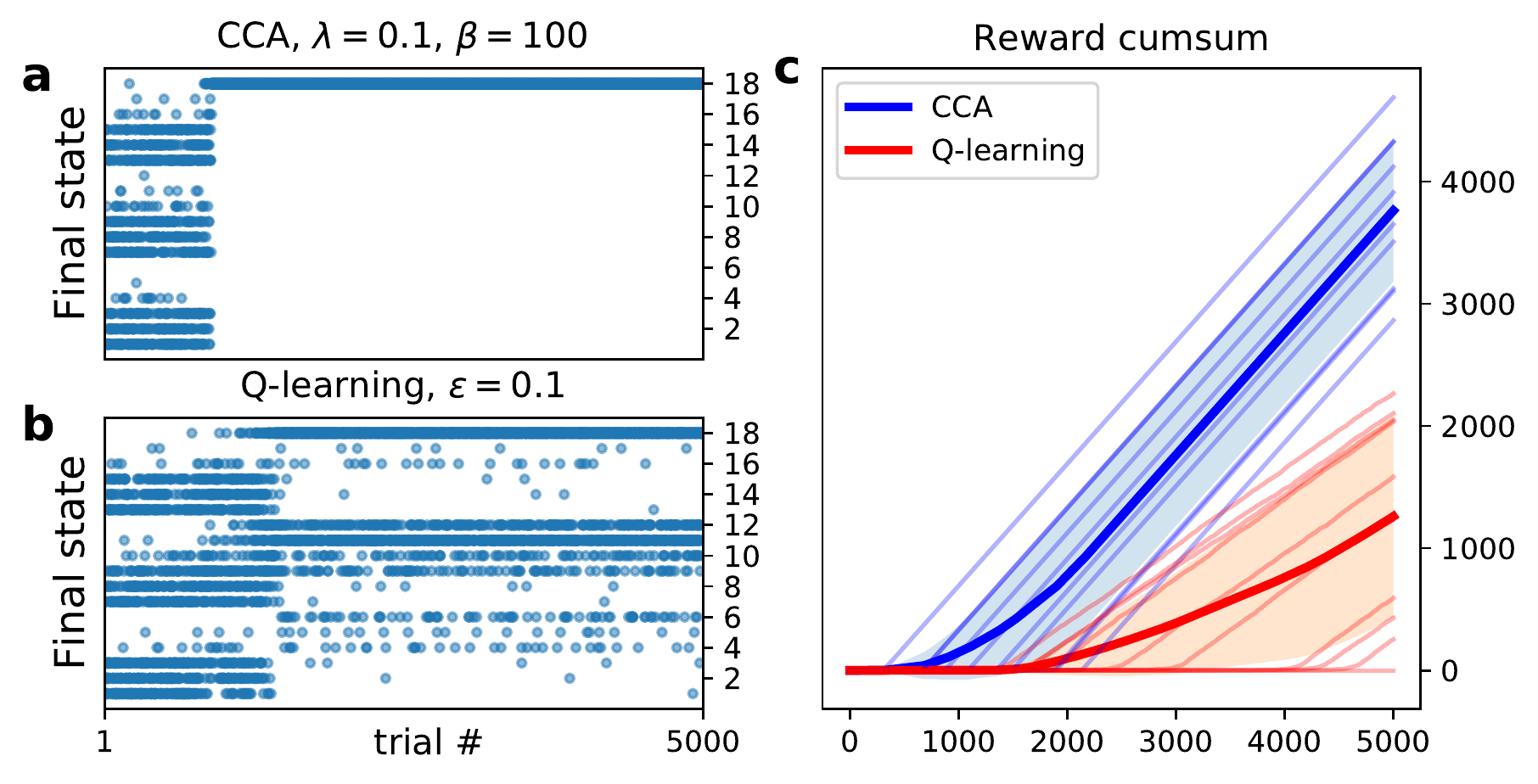}}
		\caption{{\bf Task 2 : Methods comparison (CCA ($\lambda>0$) and Q-learning)}. A reward $r=1$ is provided when the agent reaches the lower-right corner ($s'=18$). {\bf a.} Final states attained during the first 5000 trials, CCA method. {\bf b.} Final states attained during the first 5000 trials, Q-learning method. {\bf c.} The CCA algorithm is compared with Q-learning epsilon-greedy update. Cumulative sum of rewards over 5000 trials, on 10 training sessions each.  $\alpha=0.03$, $\beta = 100$, $\lambda=0.1$, $\eta=0.005$, $\varepsilon=0.1$.}\label{fig:compare}
\end{figure}

Taking $\lambda>0$ implies that extrinsic rewards are now considered in the update formula. 
For illustrative purpose, a high $\beta$ value ($\beta=100$) is taken here, in order to render the agent maximally sensitive to these extrinsic rewards, allowing to make a comparison with state-of-the art off-policy/epsilon greedy method (figure \ref{fig:compare}). With that setting, any reward encountered tends to dominate the initial exploration drive, providing a firm tendency toward an reward-seeking policy. In contrast, a pure random/$\varepsilon$-greedy exploration method would only reach the rewarding state ``by chance'', at variable delay. 
This illustrates the capability of our method to both provide an efficient exploration method when the rewards are sparse, and follow an efficient reward-seeking strategy when positive rewards are obtained.




\section{Continuous control}

Then we scale up the model towards more challenging tasks, with continuous states and actions.  Our method is tested over several benchmark environments, as provided by the ``Gym'' suite \cite{brockman2016openai}. Our implementation is moreover drawn over the ``spinning-up'' open source framework \cite{SpinningUp2018}, allowing for a direct comparison with the state of the art (here the soft actor-critic method (SAC) \cite{haarnoja2018soft}, proximal policy optimization (PPO) \cite{schulman2017proximal} and TD3 \cite{fujimoto2018addressing}). The optimization is carried out on a parameterized policy $\pi_\theta$ (said the ``actor'') and two parameterized action value function $Q_\text{R}$ and $Q_\text{ELBO}$ (said the concurrent ``critics''). 
Both the actor and the critics consist of multi-layered perceptrons, containing many parameters and organized in layered weights, over which a gradient descent is operated on losses expressed as negative objectives. 


Assuming an off-policy approach, we consider a replay buffer containing many samples of states, actions and rewards as observed from interacting with the environment $\mathcal{B}=\{...,(s,a,r,s',a'),...\}$. 
The estimation of $\rho_\pi$ is obtained from a non-parametric 
estimator, with the help of a kernel-based density estimation method \cite{pedregosa2011scikit}.
The log-occupancy is calculated on-the-fly from a sample of the replay buffer at each start of an update sequence. This sample remains quite limited in number (about 1000) in order to avoid unnecessary computer overload. 




	\begin{algorithm}
	{\small
		\caption{Actor-Critic Concurrent Credit Assignment}\label{algo:MaCAO}
		\begin{algorithmic}
			\REQUIRE{$\pi_\theta$ (actor), $Q_\text{R}$, $Q_\text{ELBO}$ (critic), $\mathcal{B}$ (replay buffer), $\beta$, $\gamma$, $\lambda$ (hyperparameters)}		
			\WHILE{number of trials not exceeded}
			\STATE initialize the environment
			\WHILE{trial not terminated}
			\STATE observe $s$
			\STATE choose $a \sim \pi_\theta(A|s)$
			\STATE read $s',r$
			\STATE store $(s,a,r,s')$ in $\mathcal{B}$
			\ENDWHILE
			\IF {$\mathcal{B}$ is full enough}
			\STATE randomly sample a batch of (next) states $\{s', ...\}$ from $\mathcal{B}$.
			\STATE estimate $\tilde{\rho}$ with a nonparametric method.
			\WHILE{number of batch updates not exceeded}
			\STATE randomly sample a batch of transitions $b = \{(s,a,r,s'), ...\}$ from $\mathcal{B}$.
			\FOR {all $(s,a,r,s') \in b$}
			\STATE sample $a'\sim \pi_\theta(A|s')$ 
			\STATE calculate $L_\text{R}(s,a,r,s')$ and $L_\text{ELBO}(s,a,r,s')$
			\ENDFOR
			\STATE update the critics $Q_\text{R}$ and $Q_\text{ELBO}$ by gradient descent over the batch.
			\FOR {all $s \in b$}
			\STATE sample $a \sim \pi_\theta(A|s)$
			\STATE calculate $L_\text{act} =  -Q_\text{R}(s,a) + \frac{1}{\beta}\log \pi_\theta(a|s) - Q_\text{ELBO}(s,a)$  
			\ENDFOR
			\STATE update the actor $\pi_\theta$ by gradient descent over the batch.	
			\ENDWHILE			
			\ENDIF			
			\ENDWHILE
	\end{algorithmic}}
\end{algorithm}

Assume two parameterize $Q$-function, i.e. $Q_\text{R}$ the cumulative rewards and $Q_\text{ELBO}$ implementing the ELBO. The policy parameters follow both a matching objective $ Q_\text{R} - \frac{1}{\beta} \log \pi_\theta$, and a policy gradient over $Q_\text{ELBO}$. 
This sketch of idea implies the use Mean-Square Bellman Errors (MSBE) to update the $Q$-functions, i.e. considering $(s,a,r,s',a')$ a sample,
\begin{small}
	\begin{align}
		L_\text{R}=& \frac{\lambda}{2} (Q_\text{R}(s,a) - r - \gamma Q_\text{R}(s',a'))^2 \label{eq:L_R}\\
		L_\text{ELBO} =& \frac{1}{2}(Q_\text{ELBO}(s,a) -  \frac{(1-\gamma)}{\beta} (L_\text{R} + \log \rho_\mathcal{D}(s') ) \nonumber\\&- \gamma Q_\text{ELBO}(s',a'))^2 \label{eq:L_ELBO}
	\end{align}
\end{small}



The main lines of our implementation are provided in algorithm \ref{algo:MaCAO}, that fits the pursuit of the ELBO objective in an actor-critic setup.
It relies on a wide use replay buffers \cite{mnih2013playing} to regularize the gradient over batches that mix the samples from many different trials. 
In order to reach state-of-the art efficacy, many algorithmic improvements are included in supplement to the baseline algorithm\footnote{code freely available at https://github.com/edauce/IJCNN-CCA.}, and omitted here for conciseness. This concerns in particular the use of target Q-networks updated at slower pace \cite{mnih2013playing}, and the clipped double-Q trick  \cite{fujimoto2018addressing}.

\begin{figure}[h!]
	\centerline{\includegraphics[width=\linewidth]{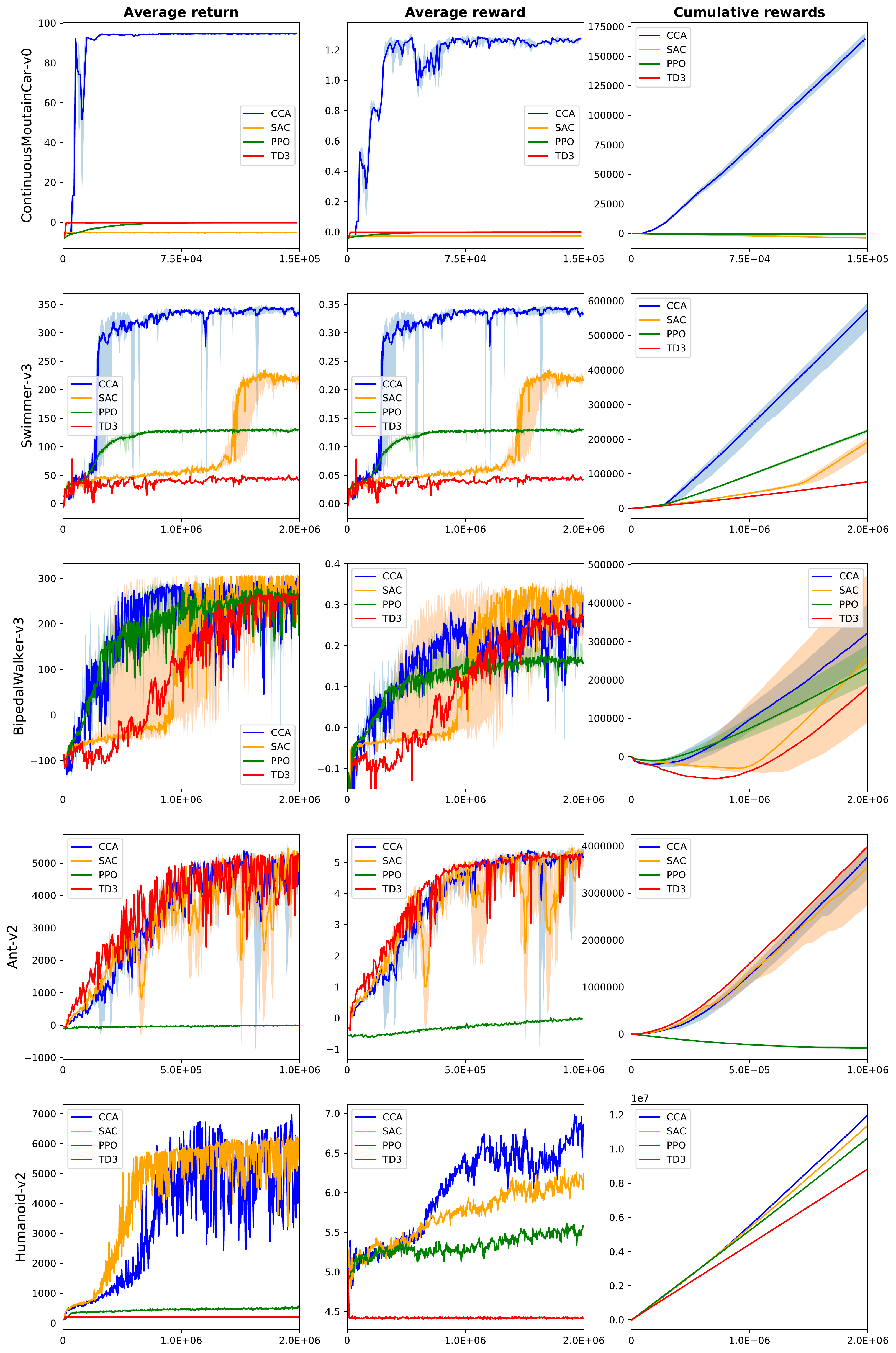}}
	\caption{
		{\bf Methods comparison.} Average episode rewards, average rewards and cumulative rewards are compared in the course of learning for the CCA (ours), SAC, PPO and TD3 frameworks, on 5 continuous state/continuous control problems. Row 1: Gym \emph{Continuous Mountain Car} problem. $\beta=10, \lambda=0.1, \gamma=0.99$, 2 hidden layers with $N=32$ neurons. 5 seeds. Row 2: Gym/MuJoCo \emph{Swimmer} problem. $\beta=30, \lambda=0.3, \gamma=0.995$, 2 hidden layers with $N=32$ neurons. 5 seeds. Row 3: Gym/Box2D \emph{Bipedal Walker}. $\beta=30, \lambda=1, \gamma=0.99$, 2 hidden layers with $N=64$ neurons. 5 seeds. Row 4: Gym/MuJoCo \emph{Ant}. $\beta=10, \lambda=0.3, \gamma=0.99$, 2 hidden layers with $N=256$ neurons. 5 seeds. Row 5: Gym/MuJoCo \emph{Humanoid}. $\beta=10, \lambda=3, \gamma=0.98$, 2 hidden layers with $N=256$ neurons. 1 seed.
	}\label{fig:method_cmp}
\end{figure}
The different setups are compared on the basis of the returns collected during training. This is expressed as average return (that is the total sum of rewards gathered at the end of an episode), the average reward (total rewards collected divided by the episode length) and cumulative rewards (the total sum of rewards collected at a given stage of the training). The width of the occupancy over the state space is not compared here, for the other frameworks are not designed to optimize it. The different environments differ in scale, difficulty and rewards density.  
All continuous problems proposed in the library provide dense rewards, that are a compound of negatively and positively weighted extrinsic informations, like the energy consumption, the speed of the agent or its elevation. The problems separate in two broad categories. A first class of problems  provides only dense rewards. A second class of problems have, in addition, a supplementary sparse reward taking the form of an ``end-of-episode'' bonus or penalty. In that case, the dense rewards may (or may not) contain relevant information with regards to the task at hand. 

From that prospect, the most adverse problem is the Continuous Mountain Car problem (first row of figure \ref{fig:method_cmp}). Here the dense rewards only rely on a (negative) energy consumption, at the exception of a +100 end-of-episode bonus obtained at the hilltop. This inevitably conducts baseline algorithms to remain stucked at the bottom of the hill, where the energy consumption is low. Only our concurrent approach, that contains an explicit incentive for widening the occupancy of the state space, has the capability to reach the most rewarding states, finally providing a policy that solves the task.  
  
The swimmer task (second row) is concerned with the development of a locomotion pattern that is swimming in a liquid medium. The reward is only the speed at which an eel-like agent manage to swim over the place (that is coordinating segments in a periodic manner). This tasks contain a local optimum that corresponds to a rower pattern that coordinates the extremal segments, and a global optimum that corresponds to a classic swimming ripple from the head toward the tail. Despite its apparent simplicity, 
only an extensive exploration such as the one provided by our approach allows to reach the optimum.   
  
The Bipedal Walker  (third row of figure \ref{fig:method_cmp}) is a problem that combines dense and sparse rewards. A negative (-100) reward is undergone when the agent falls down, and a positive (+100) reward is gained when the agent reaches the end of the track. The continual dense rewards provide an incentive for staying upright and increase the velocity. This task reveals more tricky to train than expected, and contains enough variability for the agent to develop various gaits and locomotion patterns over the course of learning. Like in the Mountain car, the problem is about reaching a final (distal) end-of-path objective, from which a strong bonus allows to ''freeze'' the behavior in a favorable locomotion pattern. Our approach shows here a clear advantage over the SAC and the TD3, and close-call with the PPO.  . 
When comparing with the SAC, the difference in the two methods is principally the \emph{time} at which a valid locomotion pattern is attained, that is about $2 \times 10^5$ interactions in the first case, and more than $10^6$ in the second. 


The fourth task, known as the ''ant'' aims at controlling the locomotion of a 4-legged agent. The state space contains a detailed account of joint angle and torque moments plus contact sensors in a 111 dimension observation vector \cite{schulman2015high}, but the control space is more reduced (8 DOFs). Here again the displacement speed is the main incentive, with a survival bonus, and an energy cost penalty. All 3 actor critic frameworks (namely CCA, SAC and CCA) are here capable to reach a decent locomotion pattern in about 400000 iterations of the dynamics, which can be considered data efficient here. No clear advantage is found here for our approach. 

Last, the humanoid task shows a large number of degrees of freedom, and the unbounded number of possible locomotion patterns often result in strange-looking final gaits. 
Only the SAC and the CCA methods allow here to reach a valid locomotion patterns in the limited number of steps considered. When looking in detail, the close-call advantage observed for the SAC algorithm on the average episode return is reversed when considering the average return. 
This apparent contradiction is explained when looking at the detailed behavior. Here, the high-speed risky locomotion patterns developed in the CCA framework result in a higher number of early failures. This is not related to a risk-seeking incentive, but is rather explained by a tendency to maintain a high diversity of behavior while pursuing the reward-guided objective, which reveals to be more risky when the balance of the body needs to be maintained over time.


\section{Discussion}

This work participates to a general trend toward the development of data models in reinforcement learning, that provide ways to help the agent toward better exploring the world. This was still largely exploited in a large family of curiosity-driven and maximum-entropy algorithms. Our contribution here is to provide a more detailed appraisal of the theoretical premises of such a construct. It is shown here to frame into a larger Bayesian/variational optimization setup, where the observations model is the variational distribution from which an evidence lower bound is maximized through gradient ascent over the policy parameters.
The general principles exposed point to the importance of this variational occupancy model that synthesizes the general distribution of the agent's observation states, over which it can act (defining a virtual ``territory''). This occupancy models is the subject of frequent updates as the exploration progresses, and that new states are undisclosed through the course of training. Making a uniform prior assumption on the occupancy results in a balance between two concurrent tendencies, namely the widening of the occupancy space and the maximization of the rewards, reminding of the classical exploration/exploitation trade-off. The consequence is a shift in the target occupancy pursued, that relaxes the constraint on fitting the initial Bellman objective.  Both are embodied in a MSBE Loss operating on two separate Q-functions in our implementation (though this is not necessary the case). Computer simulations illustrate the benefit of our conceptual developments, both in the case of sparse and dense rewards.

\bibliography{main}

\begin{thebibliography}{10}

\bibitem{10.3389/fnbot.2019.00115}
Vieri~Giuliano Santucci, Pierre-Yves Oudeyer, Andrew Barto, and Gianluca
  Baldassarre.
\newblock Editorial: Intrinsically motivated open-ended learning in autonomous
  robots.
\newblock {\em Frontiers in Neurorobotics}, 13:115, 2020.

\bibitem{sutton1998introduction}
Richard~S Sutton, Andrew~G Barto, et~al.
\newblock {\em Introduction to reinforcement learning}, volume 135.
\newblock MIT press Cambridge, 1998.

\bibitem{furmston2010variational}
Thomas Furmston and David Barber.
\newblock Variational methods for reinforcement learning.
\newblock In {\em Proceedings of the Thirteenth International Conference on
  Artificial Intelligence and Statistics}, pages 241--248. JMLR Workshop and
  Conference Proceedings, 2010.

\bibitem{levine2018reinforcement}
Sergey Levine.
\newblock Reinforcement learning and control as probabilistic inference:
  Tutorial and review.
\newblock {\em arXiv preprint arXiv:1805.00909}, 2018.

\bibitem{haarnoja2018soft}
Tuomas Haarnoja, Aurick Zhou, Pieter Abbeel, and Sergey Levine.
\newblock Soft actor-critic: Off-policy maximum entropy deep reinforcement
  learning with a stochastic actor.
\newblock In {\em International conference on machine learning}, pages
  1861--1870. PMLR, 2018.

\bibitem{abdolmaleki2018maximum}
Abbas Abdolmaleki, Jost~Tobias Springenberg, Yuval Tassa, Remi Munos, Nicolas
  Heess, and Martin Riedmiller.
\newblock Maximum a posteriori policy optimisation.
\newblock {\em arXiv preprint arXiv:1806.06920}, 2018.

\bibitem{fellows2019virel}
Matthew Fellows, Anuj Mahajan, Tim~GJ Rudner, and Shimon Whiteson.
\newblock Virel: A variational inference framework for reinforcement learning.
\newblock {\em Advances in Neural Information Processing Systems},
  32:7122--7136, 2019.

\bibitem{schmidhuber2009simple}
Jurgen Schmidhuber et~al.
\newblock Simple algorithmic theory of subjective beauty, novelty, surprise,
  interestingness, attention, curiosity, creativity, art, science, music,
  jokes.
\newblock {\em Journal of SICE}, 48(1), 2009.

\bibitem{pathak2017curiosity}
Deepak Pathak, Pulkit Agrawal, Alexei~A Efros, and Trevor Darrell.
\newblock Curiosity-driven exploration by self-supervised prediction.
\newblock In {\em International conference on machine learning}, pages
  2778--2787. PMLR, 2017.

\bibitem{oudeyer2007intrinsic}
Pierre-Yves Oudeyer, Frdric Kaplan, and Verena~V Hafner.
\newblock Intrinsic motivation systems for autonomous mental development.
\newblock {\em IEEE transactions on evolutionary computation}, 11(2):265--286,
  2007.

\bibitem{bellemare2016unifying}
Marc Bellemare, Sriram Srinivasan, Georg Ostrovski, Tom Schaul, David Saxton,
  and Remi Munos.
\newblock Unifying count-based exploration and intrinsic motivation.
\newblock {\em Advances in neural information processing systems},
  29:1471--1479, 2016.

\bibitem{tang2017exploration}
Haoran Tang, Rein Houthooft, Davis Foote, Adam Stooke, Xi~Chen, Yan Duan, John
  Schulman, Filip De~Turck, and Pieter Abbeel.
\newblock \# exploration: A study of count-based exploration for deep
  reinforcement learning.
\newblock In {\em 31st Conference on Neural Information Processing Systems
  (NIPS)}, volume~30, pages 1--18, 2017.

\bibitem{eysenbach2019if}
Benjamin Eysenbach and Sergey Levine.
\newblock If maxent rl is the answer, what is the question?
\newblock {\em arXiv preprint arXiv:1910.01913}, 2019.

\bibitem{hazan2019provably}
Elad Hazan, Sham Kakade, Karan Singh, and Abby Van~Soest.
\newblock Provably efficient maximum entropy exploration.
\newblock In {\em International Conference on Machine Learning}, pages
  2681--2691. PMLR, 2019.

\bibitem{lee2019efficient}
Lisa Lee, Benjamin Eysenbach, Emilio Parisotto, Eric Xing, Sergey Levine, and
  Ruslan Salakhutdinov.
\newblock Efficient exploration via state marginal matching.
\newblock {\em arXiv preprint arXiv:1906.05274}, 2019.

\bibitem{sabes1996reinforcement}
Philip~N Sabes and Michael~I Jordan.
\newblock Reinforcement learning by probability matching.
\newblock {\em Advances in neural information processing systems}, pages
  1080--1086, 1996.

\bibitem{herrnstein1961relative}
Richard~J Herrnstein.
\newblock Relative and absolute strength of response as a function of frequency
  of reinforcement.
\newblock {\em Journal of the experimental analysis of behavior}, 4(3):267,
  1961.

\bibitem{bellman1966dynamic}
Richard Bellman.
\newblock Dynamic programming.
\newblock {\em Science}, 153(3731):34--37, 1966.

\bibitem{o2020making}
Brendan O'Donoghue, Ian Osband, and Catalin Ionescu.
\newblock Making sense of reinforcement learning and probabilistic inference.
\newblock {\em arXiv preprint arXiv:2001.00805}, 2020.

\bibitem{dayan1993improving}
Peter Dayan.
\newblock Improving generalization for temporal difference learning: The
  successor representation.
\newblock {\em Neural Computation}, 5(4):613--624, 1993.

\bibitem{kingma2013auto}
Diederik~P Kingma and Max Welling.
\newblock Auto-encoding variational bayes.
\newblock {\em arXiv preprint arXiv:1312.6114}, 2013.

\bibitem{eysenbach2018diversity}
Benjamin Eysenbach, Abhishek Gupta, Julian Ibarz, and Sergey Levine.
\newblock Diversity is all you need: Learning skills without a reward function.
\newblock {\em arXiv preprint arXiv:1802.06070}, 2018.

\bibitem{mohamed2015variational}
Shakir Mohamed and Danilo~Jimenez Rezende.
\newblock Variational information maximisation for intrinsically motivated
  reinforcement learning.
\newblock {\em arXiv preprint arXiv:1509.08731}, 2015.

\bibitem{houthooft2016vime}
Rein Houthooft, Xi~Chen, Yan Duan, John Schulman, Filip De~Turck, and Pieter
  Abbeel.
\newblock Vime: Variational information maximizing exploration.
\newblock {\em arXiv preprint arXiv:1605.09674}, 2016.

\bibitem{williams1992simple}
Ronald~J Williams.
\newblock Simple statistical gradient-following algorithms for connectionist
  reinforcement learning.
\newblock {\em Machine learning}, 8(3):229--256, 1992.

\bibitem{puterman2014markov}
Martin~L Puterman.
\newblock {\em Markov decision processes: discrete stochastic dynamic
  programming}.
\newblock John Wiley \& Sons, 2014.

\bibitem{ho2016generative}
Jonathan Ho and Stefano Ermon.
\newblock Generative adversarial imitation learning.
\newblock {\em Advances in neural information processing systems},
  29:4565--4573, 2016.

\bibitem{schulman2017proximal}
John Schulman, Filip Wolski, Prafulla Dhariwal, Alec Radford, and Oleg Klimov.
\newblock Proximal policy optimization algorithms.
\newblock {\em arXiv preprint arXiv:1707.06347}, 2017.

\bibitem{fujimoto2018addressing}
Scott Fujimoto, Herke Hoof, and David Meger.
\newblock Addressing function approximation error in actor-critic methods.
\newblock In {\em International Conference on Machine Learning}, pages
  1587--1596. PMLR, 2018.

\bibitem{brockman2016openai}
Greg Brockman, Vicki Cheung, Ludwig Pettersson, Jonas Schneider, John Schulman,
  Jie Tang, and Wojciech Zaremba.
\newblock Openai gym.
\newblock {\em arXiv preprint arXiv:1606.01540}, 2016.

\bibitem{SpinningUp2018}
Joshua Achiam.
\newblock {Spinning Up in Deep Reinforcement Learning}.
\newblock 2018.

\bibitem{pedregosa2011scikit}
Fabian Pedregosa, Ga{\"e}l Varoquaux, Alexandre Gramfort, Vincent Michel,
  Bertrand Thirion, Olivier Grisel, Mathieu Blondel, Peter Prettenhofer, Ron
  Weiss, Vincent Dubourg, et~al.
\newblock Scikit-learn: Machine learning in python.
\newblock {\em the Journal of machine Learning research}, 12:2825--2830, 2011.

\bibitem{sutton1988learning}
Richard~S Sutton.
\newblock Learning to predict by the methods of temporal differences.
\newblock {\em Machine learning}, 3(1):9--44, 1988.

\bibitem{mnih2013playing}
Volodymyr Mnih, Koray Kavukcuoglu, David Silver, Alex Graves, Ioannis
  Antonoglou, Daan Wierstra, and Martin Riedmiller.
\newblock Playing atari with deep reinforcement learning.
\newblock {\em arXiv preprint arXiv:1312.5602}, 2013.

\bibitem{schulman2015high}
John Schulman, Philipp Moritz, Sergey Levine, Michael Jordan, and Pieter
  Abbeel.
\newblock High-dimensional continuous control using generalized advantage
  estimation.
\newblock {\em arXiv preprint arXiv:1506.02438}, 2015.

\end{thebibliography}
\bibliographystyle{unsrt}

\end{document}